\documentclass{article}

\usepackage[english]{babel}

\usepackage[letterpaper,top=2cm,bottom=2cm,left=3cm,right=3cm,marginparwidth=1.75cm]{geometry}

\usepackage{algorithm2e}
\usepackage{algpseudocode}
\usepackage{booktabs}
\usepackage{multirow}
\usepackage{amsmath}
\usepackage{amsthm}
\usepackage{graphicx}
\usepackage[colorlinks=true, allcolors=blue]{hyperref}
\usepackage{flushend, cuted}
\usepackage[title]{appendix}
\usepackage{chngcntr}
\usepackage{apptools}
\AtAppendix{\counterwithin{lemma}{section}}

\usepackage{amsfonts}
\usepackage{bbm}

\usepackage{amssymb}
\usepackage{float}
\usepackage{color,amsmath,amssymb, amsfonts,amstext,amsthm}
\usepackage{xcolor}
\usepackage{epsfig, graphicx, graphics}
\usepackage{longtable}
\usepackage{cite}
\usepackage{subfigure}

\usepackage{caption}
\usepackage{bm}
\usepackage{epstopdf}

\usepackage{color, amsmath,amssymb, amsfonts, amstext,latexsym}

\usepackage{amssymb, epsfig, amssymb, latexsym}

\usepackage{longtable}

\usepackage{authblk}

\usepackage[mathscr]{euscript}

\usepackage{color}

\renewcommand{\phi}{\varphi}

\title{Stochastic parameter reduced-order model based on hybrid machine learning approaches}

\author[a]{Cheng Fang \thanks{fangcheng1@hust.edu.cn}}
\author[b]{Jinqiao Duan \thanks{Corresponding author: duan@gbu.edu.cn}} 
\affil[a]{School of Mathematics and Statistics \& Center for Mathematical Sciences, Huazhong University of Science and Technology, Wuhan, Hubei 430074, China}
\affil[b]{Department of Mathematics and Department of Physics, Great Bay University, Dongguan, Guangdong 523000, China}

\date{March 23, 2024}

\begin{document}
\maketitle

\begin{abstract}
Establishing appropriate mathematical models for complex systems in natural phenomena not only helps deepen our understanding of nature but can also be used for state estimation and prediction. However, the extreme complexity of natural phenomena makes it extremely challenging to develop full-order models (FOMs) and apply them to studying many quantities of interest. In contrast, appropriate reduced-order models (ROMs) are favored due to their high computational efficiency and ability to describe the key dynamics and statistical characteristics of natural phenomena. Taking the viscous Burgers equation as an example, this paper constructs a Convolutional Autoencoder-Reservoir Computing-Normalizing Flow algorithm framework, where the Convolutional Autoencoder is used to construct latent space representations, and the Reservoir Computing-Normalizing Flow framework is used to characterize the evolution of latent state variables. In this way, a data-driven stochastic parameter reduced-order model is constructed to describe the complex system and its dynamic behavior.
\end{abstract}

\section{Introduction}

Perfect full-order mathematical models do not exist in practical applications due to insufficient physical understanding or computational resolution. Even if a nearly perfect full-order model can be obtained, it often has a high dimension and a complex model structure. For instance, simulating complex systems described by nonlinear partial differential equations requires high fidelity, which leads to significant computational costs. Reduced-order models (ROMs) are a method to reduce computational costs by simplifying the complexity of systems. They aim to find a low-dimensional representation of the original system while preserving its key dynamic characteristics. This approach is particularly suitable for scenarios that require fast responses or many simulations, such as in control system design, optimization algorithms, or uncertainty quantification.

Classical reduced-order models utilize appropriate dimension reduction tools to describe quantities of interest, coarse-grained representations, macro-scale behaviors, or latent state spaces. A systematic and commonly used approach involves projecting the original complex nonlinear system onto the first few large-scale modes based on the Galerkin Proper Orthogonal Decomposition\cite{holmes2012turbulence} (POD) or onto basis functions constructed by other methods such as Diffusion Maps\cite{dietrich2023learning, nadler2006diffusion, dsilva2016data, dsilva2018parsimonious} or Dynamic Mode Decomposition\cite{schmid2010dynamic}. In recent years, machine learning has received significant attention for the development of ROMs and the acceleration of complex system simulations. A prominent research approach involves utilizing Autoencoders for system dimension reduction and reconstruction, along with Recurrent Neural Networks (RNNs) to evolve the low-dimensional dynamics, as exemplified in references\cite{maulik2021reduced, lin2023online, vlachas2022multiscale, kivcic2023adaptive}.

In this paper, the viscous Burgers equation is taken as an example, which is a nonlinear partial differential equation used to simulate the propagation and reflection of advecting shocks. Leveraging Convolutional Autoencoders\cite{masci2011stacked} and the Reservoir Computing-Normalizing Flow model\cite{fang2023reservoir}, we construct a stochastic parameter reduced-order model. The focus of this paper lies in the generalization ability of this surrogate model for interpolation and extrapolation of fluid viscosity parameters in the viscous Burgers equation. Additionally, the model offers advantages such as being mesh-free, having a low-dimensional latent space, fast training speeds, and a concise model structure.

The remaining sections of the paper are organized as follows. Section \ref{5sec: prepare} introduces the problem setup and research objectives. Section \ref{5sec: SPROM} describes the fundamental frameworks of the Convolutional Autoencoder (for dimension reduction and reconstruction) and the parametric Reservoir Computing-Normalizing Flow model (for the evolution of latent variables). Subsequently, these two components are combined to construct the novel stochastic parameter reduced-order model. Section \ref{5sec: experiment} focuses on investigating the generalization ability of the model for interpolation and extrapolation of fluid viscosity parameters in the viscous Burgers equation. Finally, a summary of the paper is provided.

\section{Problem Setting}\label{5sec: prepare}
In the following, scalars are indicated by unbolded letters, and multidimensional vectors or random variables are indicated by bolded letters.

Considering the following viscous Burgers equation with a Dirichlet boundary condition in $\mathbb{R}$, which is expressed as
\begin{eqnarray} \label{5eq: burgers}
\begin{aligned}
&\frac{\partial u}{\partial t}+u\frac{\partial u}{\partial x}=\nu\frac{\partial^2u}{\partial x^2},\\
u(x,0)=u_0,\quad &x\in[0,l],\quad u(0,t)=u(l,t)=0.
\end{aligned}
\end{eqnarray}
where $t \in \mathbb{R}^+$, the state variable $u(x,t) \in \mathbb{R}$ represents the velocity, $x$ and $t$ are the spatial and temporal variables, respectively, and $\nu$ is the fluid viscosity. Even if the initial conditions are smooth and the fluid viscosity $\nu$ is small enough due to advection-dominant behavior, the partial differential equation (\ref{5eq: burgers}) can produce discontinuous solutions. At high fluid viscosity $\nu$, the equation exhibits a stronger dissipative property, while at low fluid viscosity $\nu$, advective behavior is more significant. The Reynolds number is defined as $Re = 1/\nu$. As the value of the Reynolds number $Re$ increases, the contours of the solution become sharper. Consider the following initial conditions:
\begin{eqnarray} \label{5eq: initial}
u(x,0)=\frac x{1+\sqrt{\frac1{t_0}}\exp\left(Re\ \frac{x^2}4\right)},
\end{eqnarray}
Further set $l = 1$, maximum time $t_{max} = 2$. Then the analytic solution of the partial differential equation (\ref{5eq: burgers}) exists, which can be expressed as
\begin{eqnarray} \label{5eq: analy}
u(x,t)=\frac{\frac x{t+1}}{1+\sqrt{\frac{t+1}{t_0}}\exp\left(Re\ \frac{x^2}{4t+4}\right)},
\end{eqnarray}
where $t_0 = \exp (Re/8)$.

By projecting the partial differential equation (\ref{5eq: burgers}) onto a given set of basis functions $\{\varphi_1,\varphi_2,\ldots,\varphi_K\}$, we can obtain: 
\begin{eqnarray} \label{5eq: inner}
(\partial_t u,\varphi_k)+(\nu\partial_xu,\partial_x\varphi_k)+(u u_x,\varphi_k) = 0,\quad k=1,2,\ldots,K.
\end{eqnarray}
where the symbol $(\cdot,\cdot)$ denotes the inner product. It is possible to get an approximate solution:
\begin{eqnarray}\label{5eq: approx solution}
u_d = \sum_{l = 1}^d y_l(t)\phi_l(x).
\end{eqnarray}
Here, $d$ represents the dimension of the reduced space. The time-varying reduced-order model coefficient $\bm{y} = [y_1(t),\ldots, y_d(t)]$ can be obtained by substituting $u$ with $u_d$ in equation (\ref{5eq: inner}):
\begin{eqnarray}\label{5eq: rom coefficient}
dy_k=\sum_{l=1}^d A_{kl} y_l dt+\sum_{l=1}^d \sum_{m=1}^d B_{lmk} y_l y_mdt,\quad k=1,\cdots,d,
\end{eqnarray}
where A is a $d \times d$ matrix and B is a $(d,d,d)$ tensor, defined as follows: for $k = 1,\ldots,d$, $l = 1,\ldots,d$, $m = 1,\ldots,d$,
\begin{eqnarray*}
A_{kl}=-\nu\left(\frac{\partial\varphi_l}{\partial x},\frac{\partial\varphi_k}{\partial x}\right),\quad B_{lmk}=-\left(\varphi_l\frac{\partial\varphi_m}{\partial x},\varphi_k\right).
\end{eqnarray*}
Considering the truncation error in the approximate solution $u_d$, we introduce additive noise and replace the reduced-order model coefficient $\bm{y}$ with a random vector represented by the capital letter $\bm{Y}$. We can further simplify equation~(\ref{5eq: rom coefficient}) as:
\begin{eqnarray} \label{5eq: large sde}
d \bm{Y}_t = \bm{b}(\bm{Y}_t) dt + \bm{g}d\bm{B}_t,
\end{eqnarray}
where the drift coefficient $\bm{b}(\bm{Y}_t)\in \mathbb{R}^d$ and $\bm{g}$ is a $d\times d$ real diagonal matrix with positive constant elements on the diagonal. The symbol $\bm{B}_t$ represents a Brownian motion taking values in the Euclidean space $\mathbb{R}^d$. A common approach for choosing basis functions is to use Fourier sine functions. Furthermore, basis functions can be constructed based on data characteristics using data-based construction techniques like Diffusion Maps and Proper Orthogonal Decomposition.

Based on the above description, this paper focuses on exploring the following two questions: (1) Whether the Reservoir Computing-Normalizing Flow model can serve as a surrogate model for the reduced-order model equation~(\ref{5eq: large sde}) using appropriate dimension reduction tools. (2) Since fluid viscosity $\nu$ or the Reynolds number $Re$ can affect the solution of the partial differential equation and the shape of the shock wave, it is necessary to evaluate the simulation performance of the model for parameters $Re$ that are not present in the training data, specifically the interpolation and extrapolation generalization capabilities of the surrogate model.

\section{Stochastic Parameter Reduced-Order Model}\label{5sec: SPROM}
According to the introduction in Section \ref{5sec: prepare}, constructing an appropriate reduced-order model mainly involves two aspects: constructing a latent space representation through dimension reduction tools and describing the evolution of latent state variables. In this section, a novel stochastic parameter reduced-order model is proposed. Convolutional Autoencoder is introduced in Section~\ref{5sec: CAE} as a data-driven dimension reduction tool, and the evolution of latent low-dimensional state variables is characterized using the parametric Reservoir Computing-Normalizing Flow model described in Section \ref{5sec: RC-NF}.

\subsection{Dimension Reduction: Convolutional Autoencoder} \label{5sec: CAE}
Unlike traditional methods that employ Fourier sine functions or Proper Orthogonal Decomposition to construct basis functions and achieve dimension reduction, neural network-based dimension reduction tools can more effectively obtain lower-dimensional latent space representations\cite{maulik2021reduced}. This approach can be viewed as a way to implicitly construct basis functions and explicitly obtain latent spaces. In this section, we will introduce and apply the Convolutional Autoencoder (CAE) to achieve dimension reduction.

Autoencoder (AE) is an unsupervised neural network that first compresses the input into a latent state space to obtain low-dimensional variables, and then reconstructs the output using these low-dimensional variables. It consists of two main components:

\begin{itemize}
\item The Encoder, which maps the input $\bm{u} \in \mathbb{R}^K$ to a latent space representation $\bm{Y} \in \mathbb{R}^d$, where $d \ll K$. This mapping is denoted as $\bm{F}: \mathbb{R}^K \rightarrow \mathbb{R}^d$, and the latent state variable $\bm{Y} = \bm{F}(\bm{u})$;

\item The Decoder, which maps the latent space representation $\bm{Y}$ back to the Euclidean space $\mathbb{R}^K$ to reconstruct the high-dimensional state variable $\bm{u}$. This mapping is denoted as $\bm{G}: \mathbb{R}^d \rightarrow \mathbb{R}^K$, and the estimated value $\bm{\hat{u}} = \bm{G}(\bm{Y})$.
\end{itemize}
Given the data $\bm{u}$, Autoencoder learns the low-dimensional representation $\bm{Y}$ and the mappings $\bm{F}$ and $\bm{G}$ by minimizing the reconstruction error:
\begin{eqnarray} \label{5eq: ae loss}
\begin{aligned}
\mathcal{L}_{AE} &= \sum_m \| \bm{u}^m - \bm{G}[\bm{F}(\bm{u}^m)] \|^2,\\
\bm{F},\bm{G} &= \arg \min_{\bm{F},\bm{G}} \| \bm{u}^m - \bm{G}[\bm{F}(\bm{u}^m)] \|^2.
\end{aligned}
\end{eqnarray}
where $m$ represents different samples. After training, the output of the encoder can be regarded as the latent space representation of the input data. Each scalar component of this representation can be denoted as $Y_i$, where $i = 1,\ldots,d$, and $d$ is the width of the bottleneck layer.

In this paper, we employ the Convolutional Autoencoder (CAE), which utilizes convolutional layers. Unlike fully connected layers that learn matrices mapping inputs to outputs within that layer, convolutional layers learn a set of filters $\{\bm{f}_i\}$ to extract features. Each filter $\bm{f}_i$ performs a convolution operation with a local patch of the input. The one-dimensional convolution of a filter $m_{f}$ with width $m_{f}$ and a local patch $\bm{p}$ is defined as $\bm{f}*\bm{p} :=\sum_k f_k p_k$, for $k = 1,\ldots,m_f$. Consider a one-dimensional convolutional layer with a set of filters, each having a width of $m_{f_i}$. The output neurons of each layer are composed of convolutions between the filter $\bm{f}_i$ and local patches of size $m_{f_i}$ from the input neurons. More specifically, for a typical one-dimensional convolutional layer, the output neuron corresponding to the $i$-th filter and the $j$-th local patch is given by $Y_{ij}=\sigma(\bm{f}_i*\bm{p}_j+b_i)$, where $\sigma$ is the activation function and $b_i$ is a scalar bias term. As the index $j$ increases, the local patches move along the input with stride $s$. For example, defining a one-dimensional convolutional layer with a filter $\bm{f}_0$ having a width of $m_{f} = 3$ and a stride of $s=1$, the computation of $y_{0j}$ involves the convolution of $\bm{f}_0$ with the local patch composed of input elements $j-1, j, j+1$. To perform the convolution, zero-padding is often applied to the edges of the layer's input, a process known as zero-padding. The decoder portion utilizes deconvolutional layers to return to the original dimensions. These deconvolutional layers perform upsampling using nearest-neighbor interpolation. Two-dimensional and higher-dimensional convolutions can be defined analogously. After feature extraction in the convolutional layers, their outputs are typically passed to a pooling layer for further feature selection and information filtering. The pooling layer performs a downsampling process, where its output is a subsampled representation of the input's local patches. We employ a max-pooling layer, where each output is composed of the maximum value within the corresponding local patch of the input.

\subsection{State Variable Evolution: Parametric Reservoir Computing-Normalizing Flow Model} \label{5sec: RC-NF}
The Convolutional Autoencoder plays a crucial role in dimension reduction and reconstruction. Then, we need to devise an appropriate framework to describe the evolution of low-dimensional state variables within the latent space. Drawing inspiration from the assumed reduced-order model (\ref{5eq: large sde}), this chapter adopts the Reservoir Computing-Normalizing Flow surrogate model\cite{fang2023reservoir} and makes suitable modifications. Specifically, we introduce an additional input (parameter $\nu$) into the Reservoir Computing-Normalizing Flow model, thereby creating a parametric Reservoir Computing-Normalizing Flow model.

Parametric Reservoir Computing\cite{nakajima2021reservoir,Jaeger2004,maass2002real} can be represented by the following structure:
\begin{eqnarray} \label{5eq: para rc}
    \begin{cases}
    \bm{r}_{t+1} = (1 - \alpha) \bm{r}_t + \alpha \tanh (A\bm{r}_t+ W_{in}^{\bm{Y}} \bm{Y}_t + W_{in}^{\nu} \nu + \bm{\zeta}),\\
    \hat{\bm{Y}}_{t+1} = W_{out} [1; \bm{r}_{t+1}]^T,
    \end{cases}
\end{eqnarray}
where the reservoir state is represented by an $N$-dimensional vector $\bm{r}_t$, and $\hat{\bm{Y}}_t$ denotes the predicted value from the Reservoir Computing. The $N \times N$ matrix $A$ is the adjacency matrix of the reservoir, $W_{in}^{\bm{Y}}$ is an $N \times d$ input matrix related to the latent state variables, and $W_{in}^{\nu}$ is an $N \times 1$ input matrix related to the parameters. The output layer, $W_{out}$, is a trainable $d \times (1 + N)$ readout matrix. The scalar $\alpha$ is defined as the leakage rate. The function $\tanh$ is applied element-wise to the vector components. The column vector $\bm{\zeta} \in \mathbb{R}^N$ represents the bias. For convenience, the initial value of the reservoir state can be set to $\bm{r}(t=0) = \bm{0}$. The loss function for parametric Reservoir Computing is 
\begin{equation} \label{eq: loss}
    \mathcal{L}_{RC} = \sum_{t= 1}^{T-1} ||\bm{\hat{Y}}_t - \bm{Y}_t||^2 + \lambda ||W_{out}||^2,
\end{equation}
where $\lambda > 0$ serves as the regularization hyperparameter, the term $\lambda ||W_{out}||^2$ is introduced to mitigate overfitting. By employing the Tikhonov transformation, commonly known as ridge regression, a closed-form solution can be derived for the aforementioned loss function (\ref{eq: loss}).
\begin{equation} \label{eq: Tikhonov}
    W_{out} = \bm{Y} \bm{R}^T (\bm{R} \bm{R}^T + \lambda \bm{I})^{-1},
\end{equation}
Where $\bm{R}$ represents the states matrix of dimension $(1+N) \times (T-1)$, it is constructed by concatenating column vectors $[1; \bm{r}_t]^T$ for $t = 1, \dots, T-1$. Similarly, the matrix $\bm{Y}$ is formed using $\bm{Y}_t$. Additionally, $\bm{I}$ denotes a $1 + N$ identity matrix.

The hyperparameters of the Reservoir Computing model include the spectral radius $\rho$ of the adjacency matrix $A$, the sampling range $\chi$ for the elements of the input matrix $W_{in}$, the leakage factor $\alpha$, the regularization parameter $\lambda$, the probability $p_A$ of a connection between two nodes in the reservoir state, and the probability $p_{W_{in}}$ of connecting input to a specific node in the reservoir. These hyperparameters can be adjusted to more flexibly control the performance of the reservoir computing model, as summarized in Table \ref{5tab: hyper}. We employ the Bayesian optimization algorithm\cite{fang2023reservoir,Griffith2019} to select the hyperparameters for parametric Reservoir Computing.

\begin{table}[!ht] 
\caption{The value range and data type of hyperparameters of parametric Reservoir Computing}
\label{5tab: hyper}
\centering
\begin{tabular}{ccc} 
\bottomrule
\textbf{Hyperparameters} & \textbf{Min - Max} & \textbf{Data type}\\ 
\hline
$\rho$ & 0.3 - 1.5 & Real number\\
$\chi$ & 0.3 - 1.5 & Real number \\
$\alpha$ & 0.05 - 1 & Real number\\
$\lambda$ & $10^{-10} - 1$ & Real number\\
$p_A$  & 0 - 1 & Real number\\
$p_{W_{in}}$  & 0 - 1 & Real number\\
\bottomrule
\end{tabular}
\end{table}
At time $t$, the single-step error between the predicted output $\hat{\bm{Y}}_t$ of the parametric Reservoir Computing model and the latent state variable $\bm{Y}_t$ is defined as $\tilde{\bm{\varepsilon}} := \bm{Y}_t - \hat{\bm{Y}}_t$. Set $\bm{p}(\tilde{\bm{\varepsilon}})$ represent the probability density of the error $\tilde{\bm{\varepsilon}}$. Based on the assumption of the reduced-order model (\ref{5eq: large sde}) and a fixed time step, we assume that the probability density $\bm{p}(\tilde{\bm{\varepsilon}})$ of the single-step error $\tilde{\bm{\varepsilon}}$ remains unchanged over time. Furthermore, for the invertible and differentiable mapping $\bm{h}_{\theta}$ in the Normalizing Flow model\cite{Kobyzev2021, NEURIPS2019_7ac71d43, papamakarios2021normalizing}, we employ the Rational-Quadratic Neural Spline Flow with Autoregressive layers (RQ-NSF (AR))\cite{NEURIPS2019_7ac71d43}. The Normalizing Flow model is trained based on the log-likelihood loss function. At time $t$, the predicted value of the parametric reservoir computing model after compensation by the Normalizing Flow is denoted as $\tilde{\bm{Y}_t}$.

\subsection{Stochastic Parameter Reduced-Order Model} \label{5sec: CAE-RC-NF}
\begin{figure}[!ht]
  \centering
  \includegraphics[width=0.85\textwidth]{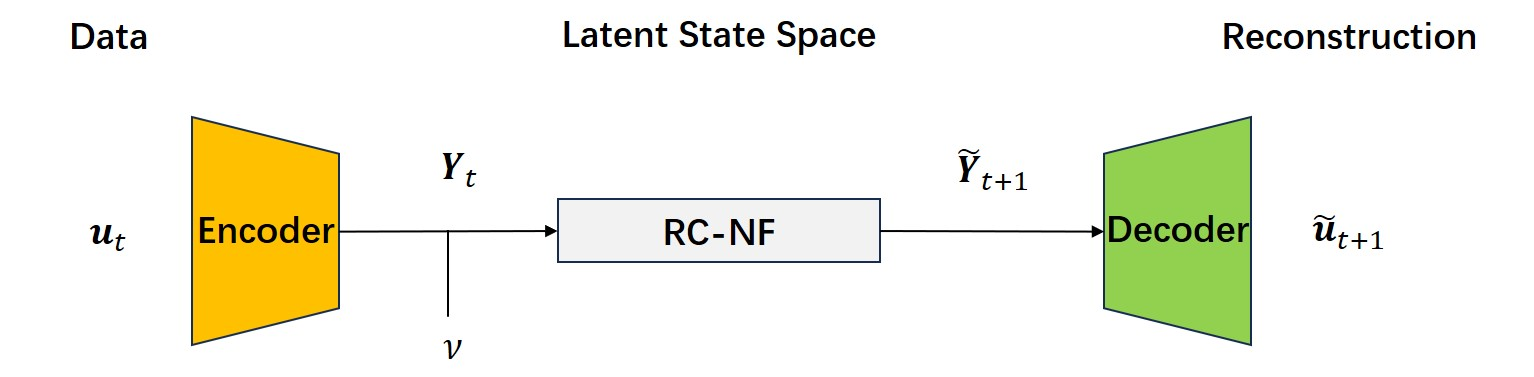}
  \caption{The flowchart of the Convolutional Autoencoder-Reservoir Computing-Normalizing Flow model. The high-dimensional state variable $\bm{u}$ is first compressed into a low-dimensional representation $\bm{Y}_t$ using an encoder. The parametric Reservoir Computing-Normalizing Flow model then characterizes the evolution of the low-dimensional variable in the latent state space. Finally, the latent state variable $\bm{Y}_t$ and its predicted value $\tilde{\bm{Y}}_t$ are reconstructed back to the original high-dimensional space through a decoder. }
  \label{5fig: CAE-RC-NF}
\end{figure}

In this paper, the Convolutional Autoencoder is utilized to replace traditional dimensionality reduction and reconstruction tools. It is combined with the parametric Reservoir Computing-Normalizing Flow model to simulate the evolution process of low-dimensional latent state variables, thereby constructing a stochastic parameter reduced-order model. This model aims to create a reduced-order surrogate model for the viscous Burgers equation (\ref{5eq: burgers}) and explore its generalization capabilities for interpolation and extrapolation of the parameter $Re$. The proposed algorithm framework is referred to as the Convolutional Autoencoder-Reservoir Computing-Normalizing Flow model (CAE-RC-NF), and its schematic diagram is shown in Figure \ref{5fig: CAE-RC-NF}. In this framework, the one-dimensional velocity variable $u$ or the corresponding high-dimensional state variable $\bm{u}$ after spatial discretization is compressed into a low-dimensional latent state space (encoder). Subsequently, the parametric Reservoir Computing-Normalizing Flow algorithm simulates the evolution of the low-dimensional variable $\bm{Y}$. Finally, the velocity variable is reconstructed using a decoder, denoted as $\tilde{u}$. Notably, to maintain training flexibility and efficiency, the two main components of the algorithm framework are trained independently, avoiding potential issues such as decreased training speed and loss function weighting that may arise from joint optimization.

\section{Experiments}\label{5sec: experiment}
This section focuses on introducing the approximation capabilities of the stochastic parameter reduced-order model for the viscous Burgers equation. The emphasis is on exploring the interpolation and extrapolation generalization abilities of the model with respect to the parameter $Re$.

\subsection{Dataset and Setting} \label{5sec: setting}
Based on the aforementioned objectives, in our numerical experiments, we constructed a dataset $\{u(x,t,Re)\} = \{u_{k,t}^{m}\}$ consisting of solutions to the partial differential equation (\ref{5eq: burgers}) under different values of the parameter $Re$. The dataset has a scale of $M \times K \times T$, where $M$ represents different values of the parameter $Re$, $K$ represents the number of equally spaced grid points for the spatial variable $x$ within the interval $[0,l]$, and $T$ represents the number of equally spaced time points for the time variable $t$ within the interval $[0,t_{max}]$. According to this setup, the training set has a size of $20 \times 128 \times 100$ with $Re$ values ranging from $100$ to $2000$ in increments of $100$. The test set has a size of $12 \times 128 \times 100$ with $Re$ values ranging from $50$ to $2450$ in increments of $200$. The spatial interval is set to $\Delta x = 1/128$, and the temporal interval is set to $\Delta t = 0.02$. It is worth noting that the $Re$ values in the test set are not present in the training set.

Next, we provide a detailed introduction to the architecture of the Convolutional Autoencoder, which aims to effectively compress and reconstruct high-dimensional state variables. The encoder employs one-dimensional convolutional layers and multiple filters with different stride lengths to obtain a low-dimensional representation of the variables. To further reduce the dimensionality of the input variables, the encoder integrates multiple convolutional layers and max-pooling layers, ultimately limiting the freedom of the latent space to only $2$ dimensions. In other words, we assume that the latent space dimension $d$ is equal to $2$. The decoder, on the other hand, is responsible for repeatedly deconvolutioning and upsampling the $2$-dimensional latent state variables to return to the full-order model.

Each one-dimensional convolutional layer utilizes filters with a width of $3$, and the subsequent max-pooling layer halves the degrees of freedom of that layer. Except for the last layers of the encoder and decoder, all other layers employ the Rectified Linear Unit (ReLU) activation function, and zero-padding is applied at the input edges of the layers to perform the convolution operation. The convolutional Autoencoder is trained using a mean squared error loss function, with a batch size of 10, and an Adam optimizer with a learning rate of 0.001. Regularization is not employed in training this model. Approximately 10\% of the training data is randomly selected to form a validation dataset, which is used for early stopping. Early stopping helps maintain the model's good generalization ability to unseen data and prevents the occurrence of overfitting.

Parametric Reservoir Computing selects 600 reservoir nodes to process the latent state variables. 10\% of the data from the training set is randomly selected to form a validation dataset. On this validation dataset, the Bayesian Optimization algorithm utilizes the mean squared error as the loss function to select the hyperparameter set for the parametric Reservoir Computing. After 100 iterations, the selected hyperparameters for the parametric Reservoir Computing are presented in Table \ref{5tab: hyper select}.
\begin{table}[!ht]
\caption{The optimal hyperparameter set selected by the Bayesian Optimization algorithm.}
\label{5tab: hyper select}
\centering
\begin{tabular}{lcccccc}
\toprule
Hyperparameter & $\rho$ & $\chi$ & $\alpha$ & $\lambda$ & $p_A$ & $p_{W_{in}}$\\
\midrule
Value & 0.1000 & 0.3332 & 1.0000 & 0.0040 & 0.9663 & 0.0165\\
\bottomrule
\end{tabular}
\end{table}

In the Normalizing Flow model, the basic transformation still utilizes the Rational Quadratic Neural Spline Flow with Autoregressive Layers (RQ-NSF (AR)). This basic transformation is composed twice to construct the final transformation $\bm{h}_\theta$. The underlying neural network in the RQ-NSF (AR) consists of a fully connected neural network with two hidden layers, each containing 8 nodes. The normalizing flow model is trained using the Adam optimizer with a learning rate of 0.005 and executed for 500 iterations.

\subsection{Results} \label{5sec: results}
\begin{figure}[!ht]
  \centering
  \includegraphics[width=0.85\textwidth]{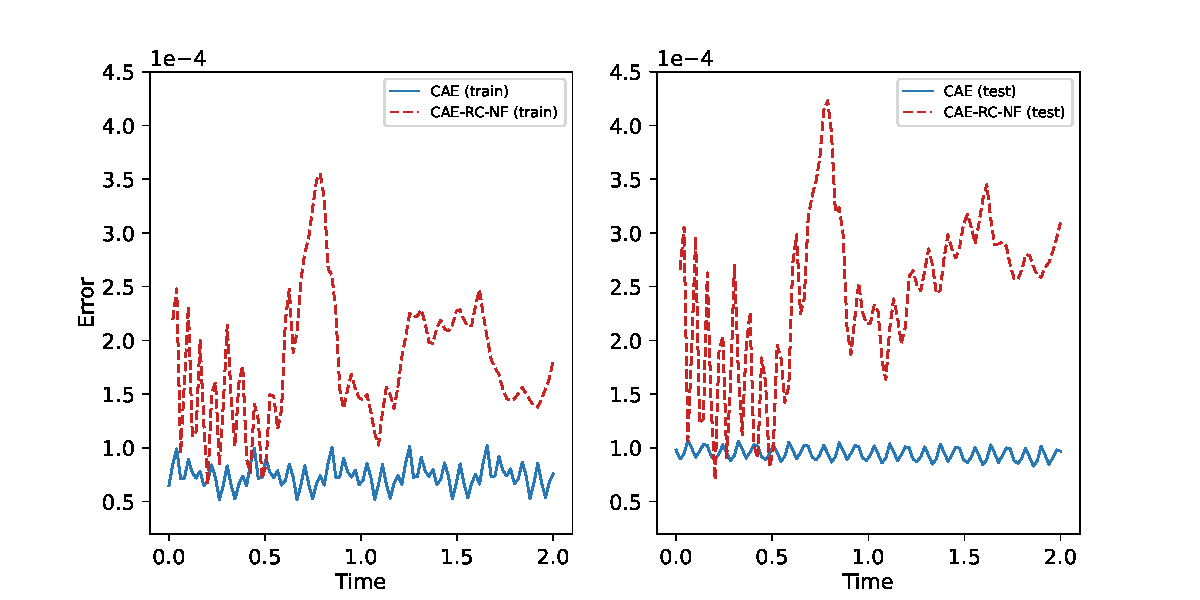}
  \caption{Reconstruction Error. The blue solid line represents the results of the Convolutional Autoencoder model, while the red dashed line represents the results of the Convolutional Autoencoder-Reservoir Computing-Normalizing Flow model. Left: Reconstruction error on the training set. Right: Reconstruction error on the test set.}
  \label{5fig: reconstruction error}
\end{figure}

First, we use reconstruction error as a criterion to demonstrate the rationality of using the Convolutional Autoencoder-Reservoir Computing-Normalizing Flow as a stochastic parameter reduction model. Specifically, at time $t \in [0, t_{max}]$, the reconstruction error of the Convolutional Autoencoder is defined as:
\begin{eqnarray}
\mathcal{L}_{CAE}^2 = \frac{1}{M\cdot K} \sum_{m=1}^{M} \sum_{k=1}^{K} \left[u_{k,t}^{m}-\bm{G}_k[\bm{F}(\bm{u}_t^{m})]\right]^2,
\end{eqnarray}
where $\bm{F}$ represents the encoder, and $\bm{G}$ represents the decoder. At time $t \in [0, t_{max}]$, the reconstruction error for the Convolutional Autoencoder-Reservoir Computing-Normalizing Flow model is defined as:
\begin{eqnarray}
\begin{aligned}
\mathcal{L}_{CAE-RC-NF}^2 &= \frac{1}{M\cdot K} \sum_{m=1}^{M} \sum_{k=1}^{K} (u_{k,t}^{m}-\tilde{u}_{k,t}^{m})^2\\
 &= \frac{1}{M\cdot K} \sum_{m=1}^{M} \sum_{k=1}^{K} \left[u_{k,t}^{m}-\bm{G}_k[\tilde{\bm{Y}}_t^{m}]\right]^2,
\end{aligned}
\end{eqnarray}
where $\tilde{\bm{Y}}_t^{m}$ represents the predicted values from the Reservoir Computing-Normalizing Flow model in the latent state space. Specifically, after training the Reservoir Computing-Normalizing Flow model, we utilize a short warm-up trajectory of length $T_{warm} = 10$ to generate a new trajectory of a specified length. This warm-up data is processed through the encoder, which is trained on (or tested with) the dataset $\{u(x,t,Re)\}$. Subsequently, the reconstruction error is calculated and presented in Figure \ref{5fig: reconstruction error}.

The results indicate that both the Convolutional Autoencoder model and the Convolutional Au-toencoder-Reservoir Computing-Normalizing Flow model exhibit relatively small reconstruction errors on the training set. However, the reconstruction errors are somewhat larger on the test set, but they remain within the same order of magnitude ($10^{-4}$). Since the error of the Convolutional Autoencoder-Reservoir Computing-Normalizing Flow model comprises both the reconstruction error of the Convolutional Autoencoder and the prediction error of the Reservoir Computing-Normalizing Flow, its overall reconstruction error is relatively larger.

\begin{figure}[!ht]
  \centering
  \includegraphics[width=0.85\textwidth]{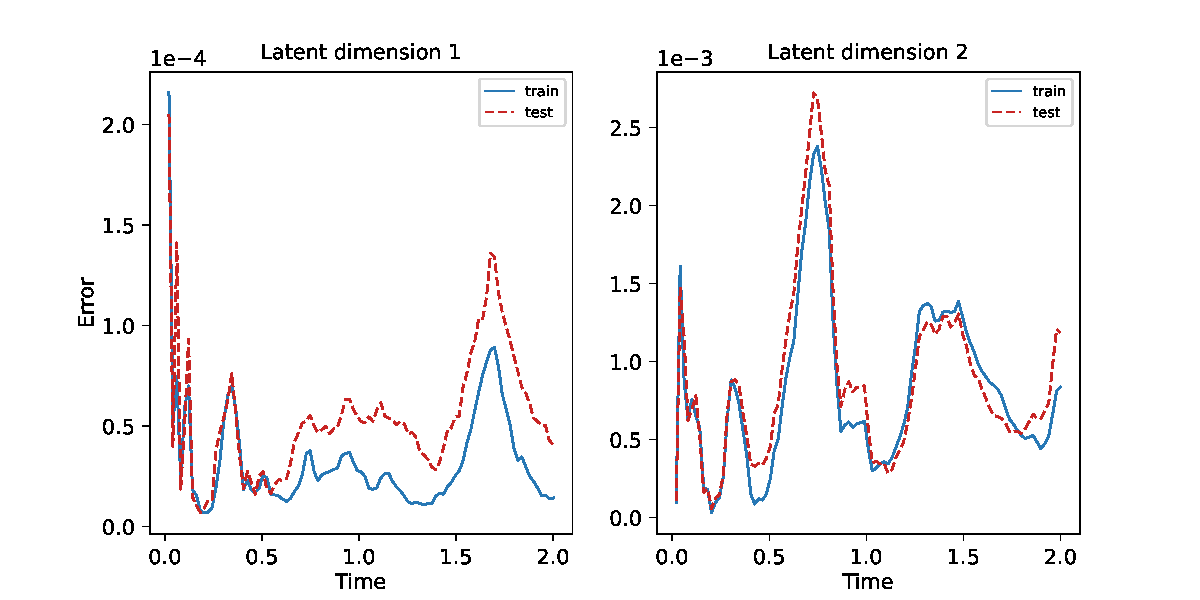}
  \caption{Prediction error of the Reservoir Computing-Normalizing Flow Model. The blue solid line represents the prediction results of the model on the training set, while the red dashed line represents the prediction results on the test set. Left: Prediction error for latent state space dimension 1. Right: Prediction error for latent state space dimension 2.}
  \label{5fig: latent error}
\end{figure}

We further calculate the prediction error of the parametric Reservoir Computing-Normalizing Flow framework in the latent state space. Specifically, at any given time $t \in (0, t_{max}]$, the prediction error of the parametric Reservoir Computing-Normalizing Flow model is defined as
\begin{eqnarray}
\mathcal{L}_{RC-NF}^2 = \frac{1}{M} \sum_{m=1}^{M} \left[\bm{Y}_{i,t}^{m}-\tilde{\bm{Y}}_{i,t}^{m}\right]^2,\quad i = 1, 2,
\end{eqnarray}
where the latent space variable $\bm{Y}$ is obtained through compression by the encoder, while $\tilde{\bm{Y}}$ represents the predicted output from the parameter reservoir computing-normalizing flow model. The results of the prediction error are presented in Figure \ref{5fig: latent error}. The prediction error of the Reservoir Computing-Normalizing Flow model remains consistent in magnitude across both the training set and the test set. Notably, due to the introduction of randomness by the Normalizing Flow, the prediction error of the parameter Reservoir Computing-Normalizing Flow model on the test set is even lower in certain periods (as evident in the right subfigure of Figure \ref{5fig: latent error}). In the following sections of this chapter, we will showcase the performance of the Convolutional Autoencoder-Reservoir Computing-Normalizing Flow model on specific test sets, highlighting its generalization capabilities for interpolating and extrapolating the parameter $Re$.

\begin{figure}[!ht]
  \centering
  \includegraphics[height=2in]{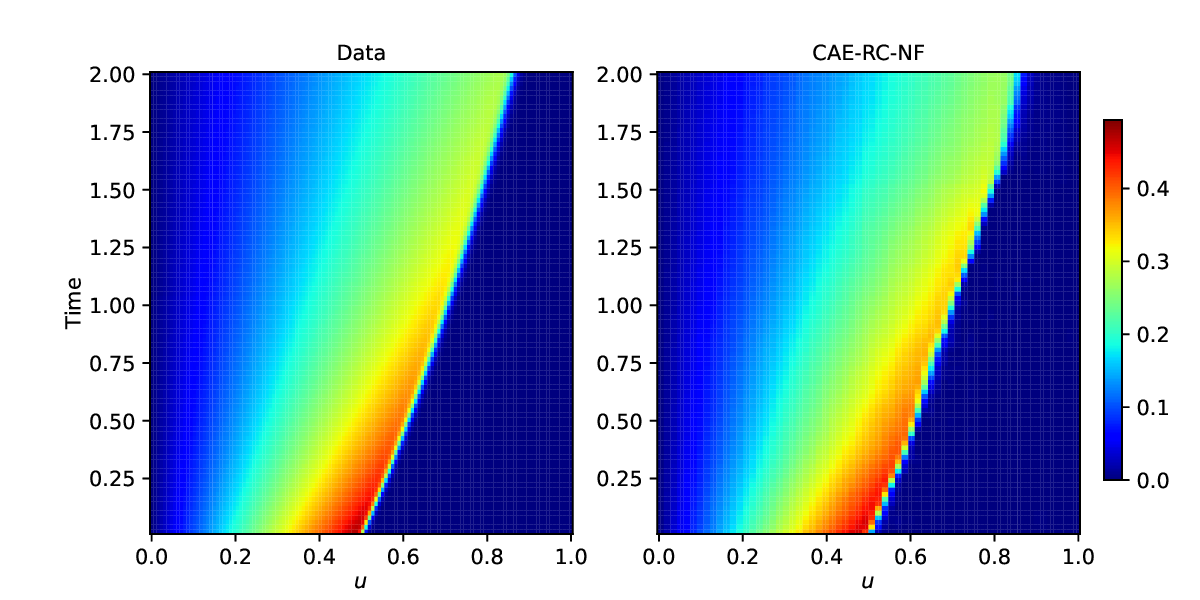}
  \includegraphics[height=2in]{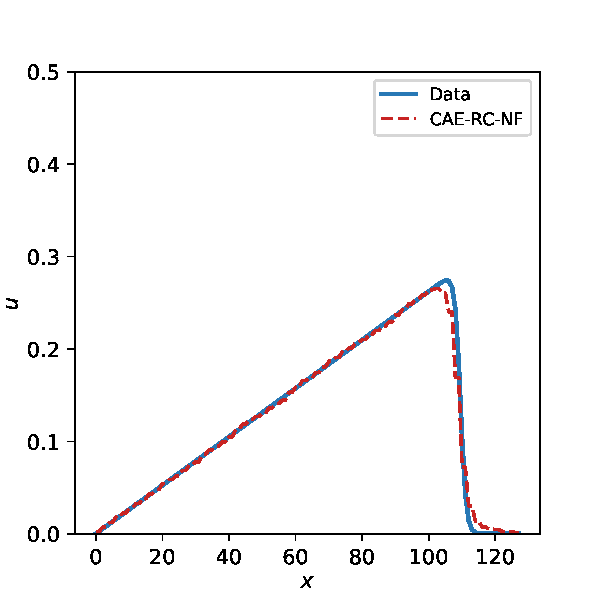}
  \makebox[\textwidth][c]{
  \includegraphics[width=0.75\textwidth]{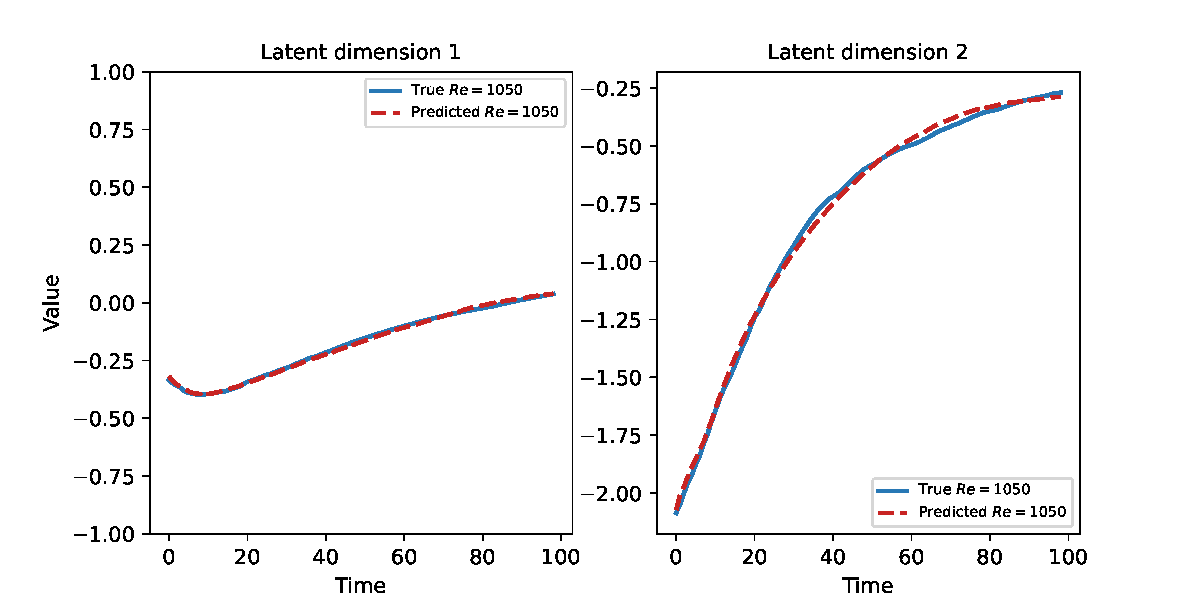}}
  \caption{Results for parameter $Re=1050$. Top Left: Data (left) and reconstruction results using the Convolutional Autoencoder-Reservoir Computing-Normalizing Flow model (right). Top Right: Data and reconstruction results at time $t = 2$, with the reference (blue solid line) and the prediction from the Convolutional Autoencoder-Reservoir Computing-Normalizing Flow model (red dashed line). Bottom: Latent state variables constructed by the encoder (blue solid line) and prediction results from the Reservoir Computing-Normalizing Flow model (red dashed line).}
  \label{5fig: re1050}
\end{figure}

\begin{figure}[!ht]
  \centering
  \includegraphics[height=2in]{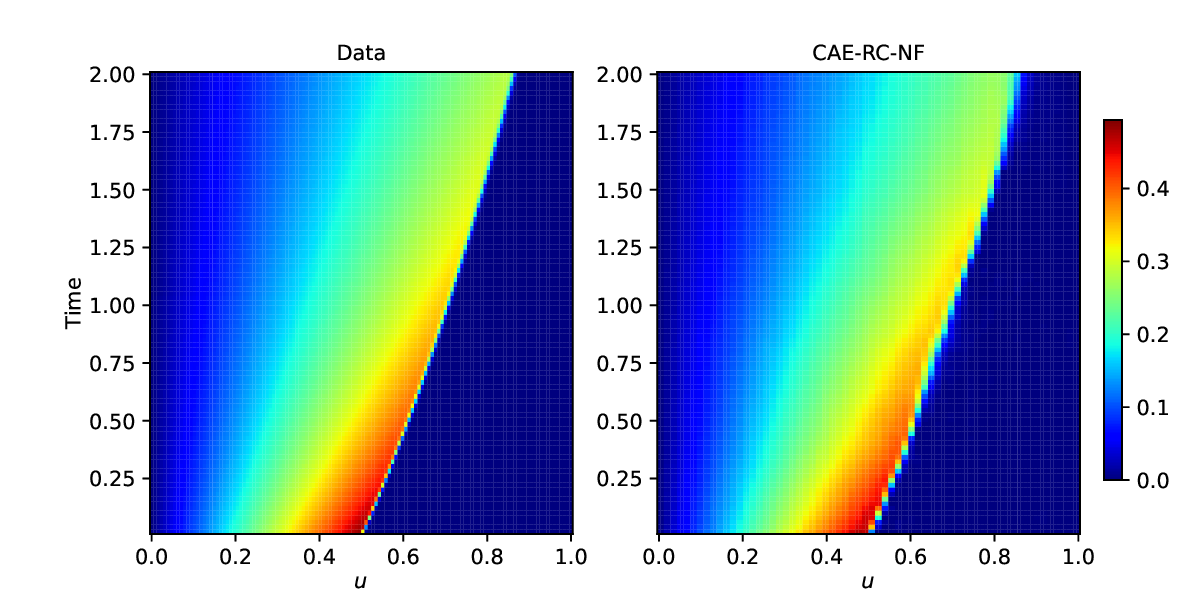}
  \includegraphics[height=2in]{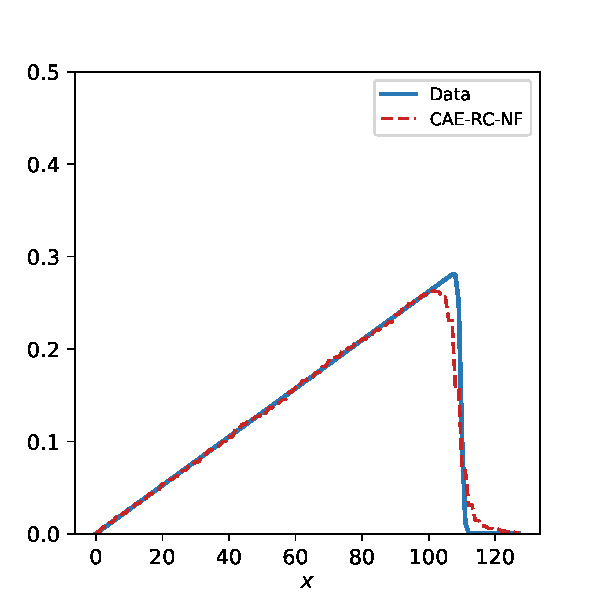}
  \makebox[\textwidth][c]{
  \includegraphics[width=0.75\textwidth]{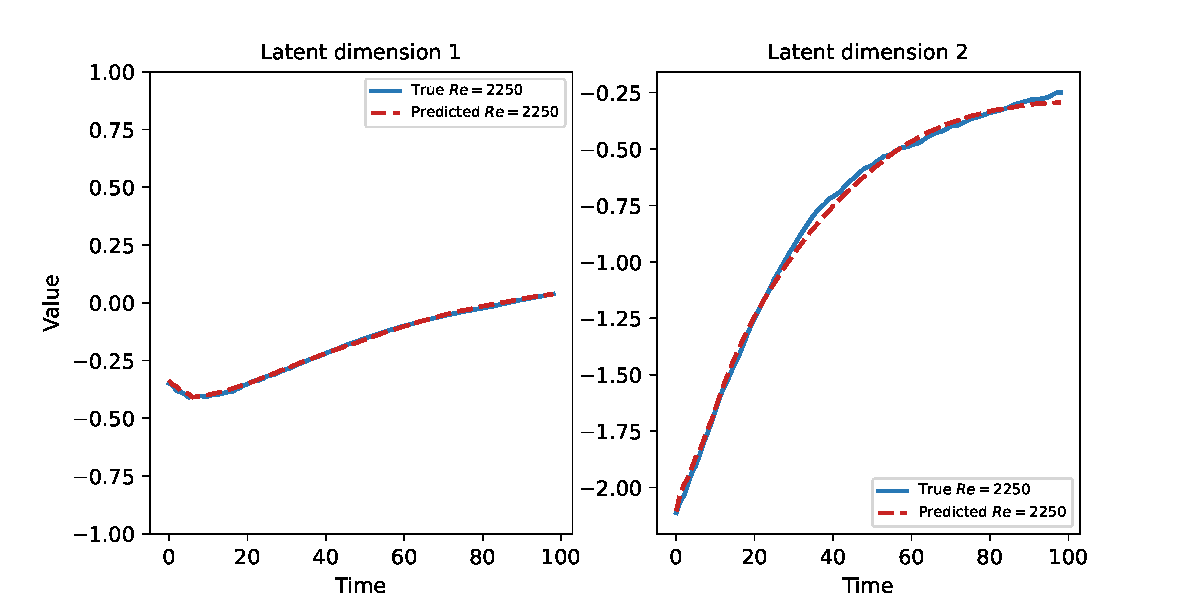}}
  \caption{Results for parameter $Re=2250$. Top Left: Data (left) and reconstruction results using the Convolutional Autoencoder-Reservoir Computing-Normalizing Flow (right). Top Right: Data and reconstruction results at time $t = 2$, with the reference (blue solid line) and the Convolutional Autoencoder-Reservoir Computing-Normalizing Flow (red dashed line). Bottom: Latent state variables constructed by the encoder (blue solid line) and prediction results from the Reservoir Computing-Normalizing Flow (red dashed line).}
  \label{5fig: re2250}
\end{figure}

We demonstrate the generalization capabilities of the Convolutional Autoencoder-Reservoir Computing-Normalizing Flow model using the examples of parameters $Re=1050$ (interpolation) and $Re=2250$ (extrapolation). Figure \ref{5fig: re1050} presents the results for $Re=1050$, including data and reconstruction results for all time instances $t \in (0,t_{max}]$ (top left subfigure), as well as data and prediction results at a specific time $t = 2$ (top right subfigure). The experimental results indicate that the proposed framework successfully reproduces the advection shock wave phenomenon in the viscous Burgers equation. The bottom subfigure of Figure~\ref{5fig: re1050} shows the prediction results of the Reservoir Computing-Normalizing Flow model in the latent state space, confirming that it can serve as an alternative model for the evolution of latent state variables. Figure \ref{5fig: re2250} displays the relevant results for $Re=2250$ and further illustrates the effectiveness of the Convolutional Autoencoder-Reservoir Computing-Normalizing Flow model. These simulation experiments fully demonstrate the model's generalization capabilities for interpolating and extrapolating the parameter $Re$, thus validating its effectiveness as a reduced-order surrogate model for the partial differential equation (\ref{5eq: burgers}). This achievement further confirms that the algorithm proposed in this chapter can serve as a reliable stochastic parameter reduced-order model, providing support for simulating and predicting complex systems.

\section{Summary}\label{5sec: summary}
By combining data-driven dimension reduction tools, specifically Convolutional Autoencoders, with the parametric Reservoir Computing-Normalizing Flow model that characterizes the evolution of latent state variables, this paper constructs a data-driven stochastic parameter reduced-order model. This model, serving as a reduced-order surrogate for partial differential equations, successfully captures the advection shock wave phenomenon described by the viscous Burgers equation. Additionally, the model boasts advantages such as being mesh-free and having a low-dimensional latent space. It further demonstrates generalization capabilities for interpolating and extrapolating the parameter $Re$. For more complex and higher-dimensional partial differential equations, the effectiveness of this framework remains to be further studied and explored.

\section*{Declaration of competing interest}
The authors declare that they have no known competing financial interests or personal relationships that could have appeared to influence the work reported in this paper.

\section*{Acknowledgements}
We would like to thank Yubin Lu for helpful discussions. This work was supported by the Fundamental Research Funds for the Central Universities, HUST: 2023JYCXJJ045.

\bibliographystyle{unsrt}
\bibliography{sample.bib}

\begin{thebibliography}{10}

\bibitem{holmes2012turbulence}
Philip Holmes.
\newblock {\em Turbulence, coherent structures, dynamical systems and symmetry}.
\newblock Cambridge university press, 2012.

\bibitem{dietrich2023learning}
Felix Dietrich, Alexei Makeev, George Kevrekidis, Nikolaos Evangelou, Tom Bertalan, Sebastian Reich, and Ioannis~G Kevrekidis.
\newblock Learning effective stochastic differential equations from microscopic simulations: Linking stochastic numerics to deep learning.
\newblock {\em Chaos: An Interdisciplinary Journal of Nonlinear Science}, 33(2), 2023.

\bibitem{nadler2006diffusion}
Boaz Nadler, St{\'e}phane Lafon, Ronald~R Coifman, and Ioannis~G Kevrekidis.
\newblock Diffusion maps, spectral clustering and reaction coordinates of dynamical systems.
\newblock {\em Applied and Computational Harmonic Analysis}, 21(1):113--127, 2006.

\bibitem{dsilva2016data}
Carmeline~J Dsilva, Ronen Talmon, C~William Gear, Ronald~R Coifman, and Ioannis~G Kevrekidis.
\newblock Data-driven reduction for a class of multiscale fast-slow stochastic dynamical systems.
\newblock {\em SIAM Journal on Applied Dynamical Systems}, 15(3):1327--1351, 2016.

\bibitem{dsilva2018parsimonious}
Carmeline~J Dsilva, Ronen Talmon, Ronald~R Coifman, and Ioannis~G Kevrekidis.
\newblock Parsimonious representation of nonlinear dynamical systems through manifold learning: A chemotaxis case study.
\newblock {\em Applied and Computational Harmonic Analysis}, 44(3):759--773, 2018.

\bibitem{schmid2010dynamic}
Peter~J Schmid.
\newblock Dynamic mode decomposition of numerical and experimental data.
\newblock {\em Journal of fluid mechanics}, 656:5--28, 2010.

\bibitem{maulik2021reduced}
Romit Maulik, Bethany Lusch, and Prasanna Balaprakash.
\newblock Reduced-order modeling of advection-dominated systems with recurrent neural networks and convolutional autoencoders.
\newblock {\em Physics of Fluids}, 33(3), 2021.

\bibitem{lin2023online}
Sen Lin, Gianmarco Mengaldo, and Romit Maulik.
\newblock Online data-driven changepoint detection for high-dimensional dynamical systems.
\newblock {\em Chaos: An Interdisciplinary Journal of Nonlinear Science}, 33(10), 2023.

\bibitem{vlachas2022multiscale}
Pantelis~R Vlachas, Georgios Arampatzis, Caroline Uhler, and Petros Koumoutsakos.
\newblock Multiscale simulations of complex systems by learning their effective dynamics.
\newblock {\em Nature Machine Intelligence}, 4(4):359--366, 2022.

\bibitem{kivcic2023adaptive}
Ivica Ki{\v{c}}i{\'c}, Pantelis~R Vlachas, Georgios Arampatzis, Michail Chatzimanolakis, Leonidas Guibas, and Petros Koumoutsakos.
\newblock Adaptive learning of effective dynamics for online modeling of complex systems.
\newblock {\em Computer Methods in Applied Mechanics and Engineering}, 415:116204, 2023.

\bibitem{masci2011stacked}
Jonathan Masci, Ueli Meier, Dan Cire{\c{s}}an, and J{\"u}rgen Schmidhuber.
\newblock Stacked convolutional auto-encoders for hierarchical feature extraction.
\newblock In {\em Artificial Neural Networks and Machine Learning--ICANN 2011: 21st International Conference on Artificial Neural Networks, Espoo, Finland, June 14-17, 2011, Proceedings, Part I 21}, pages 52--59. Springer, 2011.

\bibitem{fang2023reservoir}
Cheng Fang, Yubin Lu, Ting Gao, and Jinqiao Duan.
\newblock Reservoir computing with error correction: Long-term behaviors of stochastic dynamical systems.
\newblock {\em Physica D: Nonlinear Phenomena}, 456:133919, 2023.

\bibitem{nakajima2021reservoir}
Kohei Nakajima and Ingo Fischer.
\newblock {\em Reservoir computing: Theory, Physical Implementations, and Applications}.
\newblock Springer Singapore, 2021.

\bibitem{Jaeger2004}
Herbert Jaeger and Harald Haas.
\newblock Harnessing nonlinearity: Predicting chaotic systems and saving energy in wireless communication.
\newblock {\em Science}, 304(5667):78--80, 2004.

\bibitem{maass2002real}
Wolfgang Maass, Thomas Natschl{\"a}ger, and Henry Markram.
\newblock Real-time computing without stable states: A new framework for neural computation based on perturbations.
\newblock {\em Neural computation}, 14(11):2531--2560, 2002.

\bibitem{Griffith2019}
Aaron Griffith, Andrew Pomerance, and Daniel~J. Gauthier.
\newblock Forecasting chaotic systems with very low connectivity reservoir computers.
\newblock {\em Chaos: An Interdisciplinary Journal of Nonlinear Science}, 29(12), 12 2019.
\newblock 123108.

\bibitem{Kobyzev2021}
Ivan Kobyzev, Simon~J.D. Prince, and Marcus~A. Brubaker.
\newblock Normalizing flows: An introduction and review of current methods.
\newblock {\em IEEE Transactions on Pattern Analysis and Machine Intelligence}, 43(11):3964--3979, 2021.

\bibitem{NEURIPS2019_7ac71d43}
Conor Durkan, Artur Bekasov, Iain Murray, and George Papamakarios.
\newblock Neural spline flows.
\newblock In H.~Wallach, H.~Larochelle, A.~Beygelzimer, F.~d\textquotesingle Alch\'{e}-Buc, E.~Fox, and R.~Garnett, editors, {\em Advances in Neural Information Processing Systems}, volume~32. Curran Associates, Inc., 2019.

\bibitem{papamakarios2021normalizing}
George Papamakarios, Eric Nalisnick, Danilo~Jimenez Rezende, Shakir Mohamed, and Balaji Lakshminarayanan.
\newblock Normalizing flows for probabilistic modeling and inference.
\newblock {\em The Journal of Machine Learning Research}, 22(1):2617--2680, 2021.

\end{thebibliography}

\end{document}